\documentclass[lettersize,journal]{IEEEtran}

\usepackage{amsmath,amsfonts}
\usepackage{algorithmic}
\usepackage{algorithm}
\usepackage{graphicx} %
\usepackage{booktabs}
\usepackage{lipsum}
\usepackage{comment}
\usepackage[caption=false,font=normalsize,labelfont=sf,textfont=sf]{subfig}
\usepackage{textcomp}
\usepackage{stfloats}
\usepackage{cite}
\usepackage{verbatim}

\usepackage[T1]{fontenc}
\usepackage{array}

\usepackage[export]{adjustbox}
\usepackage[colorlinks = true,
            linkcolor = black,
            urlcolor  = black,
            citecolor = black,
            anchorcolor = black]{hyperref}
\usepackage{siunitx}
\usepackage{todonotes}
\usepackage{mathtools}
\usepackage{layouts}
\usepackage{multirow}
\usepackage{url}
\usepackage{amssymb}
\usepackage{dcolumn}
\newcommand{\etal}{~\emph{et al.}}

\newcommand{\tabbaseline}{\begin{tabular}{lcccr}
\toprule
 & Density & $a\metricName (\downarrow)$ & $N_{o}$ & Time [s] \\
\midrule
Dube\etal~\cite{dube2006} & 0.213 & 1.5 & 3.2 & \textbf{0.173} \\
GOPT~\cite{xiong2024gopt} & 0.252 & 1.4 & 8.2 & 0.357 \\
Elhedhli\etal~\cite{elhedhli2019} & 0.167 & 0.4 & 1.5 & 1178.702 \\
\algName & 0.445 & \textbf{0.2} & 6.5 & 2.990 \\
\algName~(vanilla) & \textbf{0.478} & 2.4 & 6.2 & 3.631 \\
\bottomrule
\end{tabular}
}

\newcommand{\tabsensitivityStudy}{\begin{tabular}{lcccc}
\toprule
 & Density  & $a\metricName (\downarrow)$  & $N_{o}$ & Time [s]  \\
\midrule
GPT-5 & 0.445 & \textbf{0.2} & 6.5 & 2.990 \\
GPT-5-mini & 0.433 & 0.5 & 5.6 & 3.656 \\
Claude Sonnet 4.5 & \textbf{0.478} & 1.6 & 7.1 & \textbf{2.516} \\
Gemini 3 Flash & 0.466 & 0.75 & 6.0 & 2.586 \\
\bottomrule
\end{tabular}
}

\usepackage{etoolbox}
\makeatletter
\patchcmd{\@makecaption}
  {\scshape}
  {}
  {}
  {}
\makeatletter
\patchcmd{\@makecaption}
  {\\}
  {.\ }
  {}
  {}
\makeatother
\def\tablename{Table}

\newif\ifdraft
\draftfalse %

\newif\ifnewtitle
\newtitlefalse

\ifnewtitle
\newcommand{\algName}{SafePack}
\else 
\newcommand{\algName}{iPack}
\fi

\newcommand{\metricNameFull}{Packing Consistency Score ($C$)}
\newcommand{\metricName}{C}

\begin{document}
\title{
\ifnewtitle

\else
\algName: Intuitive Bin Packing with Large Language Models
\fi
}

\ifdraft
\else
\author{Yannik Blei$^{1}$, Michael Krawez$^{1}$, Adrian Gö{\ss}$^{1}$, Devadas Vijayan Sheela$^{2}$, Tobias Jülg$^{1}$, Pierre Krack$^{1}$,\\ Florian Walter$^{3, 1}$ and Wolfram Burgard$^{1}$%
\thanks{$^{1}$University of Technology Nuremberg, $^2$Friedrich-Alexander  University of Erlangen–Nuremberg, $^{3}$Technical University of Munich}
}
\fi

\maketitle

\bstctlcite{IEEEexample:BSTcontrol}

\begin{abstract}
Robotics and automation are increasingly influential in logistics but remain largely confined to traditional warehouses. 
Additionally, while there has been a substantial body of work in robotics on bin picking~\cite{cordeiro2022bin}, packing objects and groceries has remained largely unexplored. 
However, packing items in the right order is crucial for preventing product damage, e.g., heavy objects should not be placed on top of fragile ones. 
In practice, the exact criteria for the right packing order are hard to define, particularly given the wide variety of objects typically found in stores. 
In this paper, we introduce \algName, a novel approach for bin packing. \algName~leverages language and vision foundation models for identifying objects and generating packing constraints that mimic human packing strategies. 
It employs these constraints in a mixed-integer optimization framework that computes an optimal packing scheme, which is then achieved  by a robot manipulator in an offline 3D bin packing task. A major advantage of \algName~is that it does not require any training to complete the task or handle new items. Additionally, it can be utilized to select the smallest suitable container from a given set of objects.
We extensively evaluate our approach to demonstrate its performance. 
\ifdraft
\else
We make the source code of \algName~publicly available on GitHub.\footnote{\url{https://github.com/yblei/ipack}}
\fi

\end{abstract}

\begin{IEEEkeywords}
bin packing, large language models, robot manipulation, grocery packing, mixed-integer linear programming
\end{IEEEkeywords}

\section{Introduction}

Automatic packing of objects in bins or boxes with robotic manipulators has been studied extensively, primarily in the context of warehouse automation and manufacturing \cite{pantoja2024comprehensive, amoo2024warehouse}. Past approaches focused on optimizing the packing density and object stability within the bin~\cite{agarwal2020jampacker, wang2019stable, wang2022dense}. Outside of controlled warehouses, however, item properties like weight, fragility, or object class can be crucial to determine the packing order. For example, heavy objects should not be placed on top of fragile ones, frozen groceries should not be packed on top of moisture-sensitive objects, and during a move semantic reasoning is required to order household items between removal crates. Humans can handle such packing order tasks with ease and intuitively adapt to a given context. We aim to close this gap in bin packing by introducing semantic constraints to the packing scheme, whereas we propose to use the common sense knowledge of large pretrained language models to generate human-like constraints. 

We demonstrate this concept on the case study of automated grocery packing, a domain in which the wide variety of item types renders manual specification of packing rules infeasible. Owing to the generality of the underlying LLMs, the proposed approach can be readily adapted to alternative packing domains and task-specific user preferences. Prior work on grocery store automation has predominantly addressed mobile navigation and object detection~\cite{cheng2017design,thompson2018autonomous,priya2021autonomous,dworakowski2021robot}, leaving the packing stage largely unexplored.

To address this gap, we introduce \algName, a novel approach that combines Vision Language Models (VLMs) for semantic knowledge, Vision Foundation Models (VFMs) for accurate object localization, and a Mixed-Integer Linear Programming (MILP) framework for spatial optimization.

\algName~operates in an open-vocabulary manner and does not require re-training when facing novel item classes.
Its modular design further allows seamless integration of improved foundation models as they become available. 
In addition, the proposed formulation naturally extends to related logistics settings, including selecting a minimally sized container from a predefined set, incorporating constraints for packing stability and restricting placement poses, unreachable by a robotic manipulator.

\begin{figure}
    \centering
    \includegraphics[width=0.95\linewidth]{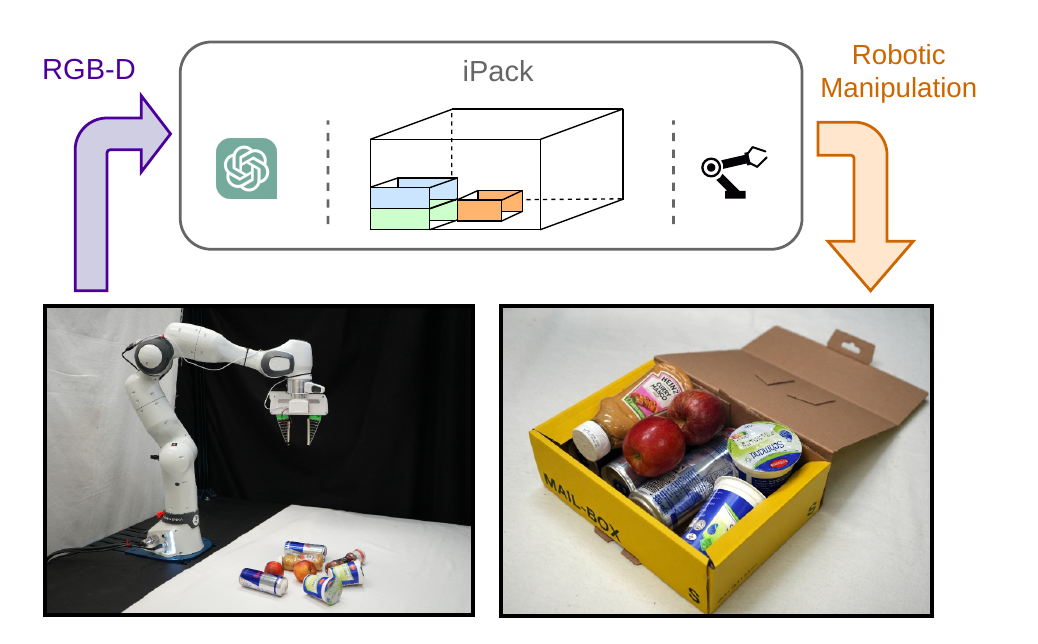}
    \caption{\algName~ensures product integrity during bin packing. It employs foundation models to identify items and packing constraints from a scene image. It then uses a mixed-integer linear programming approach to compute a spatially optimized packing scheme. A robot manipulator  places each item at its target position.}
    \label{fig:cg} 
\end{figure}

In summary, we make the following contributions:
\begin{enumerate}
    \item We propose a novel approach to open-vocabulary object packing that optimizes packing space under dynamically generated product integrity constraints.
    \item We propose the \metricNameFull, a novel data-driven metric for human-like packing.
    \item We extensively evaluate our method to demonstrate its applicability in simulation and on real robots.
    \item We release the code and the evaluation dataset, including the manipulation stack, enabling easy integration of our method in custom scenarios.
\end{enumerate}

\section{Related Work}
\begin{figure*}[ht]
    \vspace{1mm} %
    \centering
    \includegraphics[width=1\linewidth]{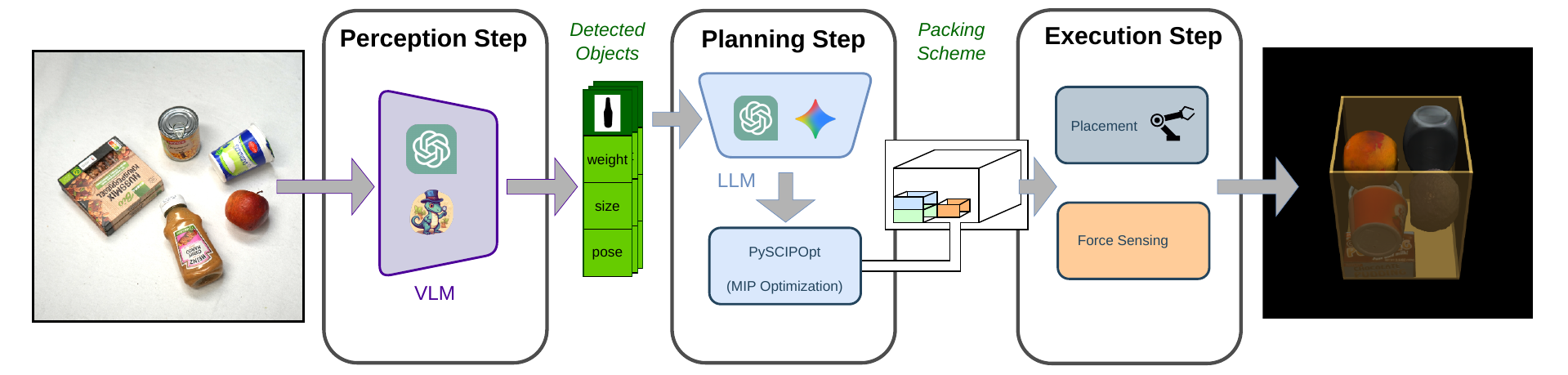}
    \caption{Overview of \algName{}: We assume a random but separated arrangement of items on the table. We employ GroundingDINO~\cite{liu2024grounding} to detect all objects, followed by SAM 2~\cite{ravi2025sam} to obtain segmentation masks. We employ a VLM to obtain estimates of the sizes and the weights of the objects. In the planning step, we utilize a large language model (LLM) to determine product integrity constraints. We apply a mixed-integer linear programming approach \cite{Maher2016} to optimize the packing space. In the final execution step, we perform the manipulation tasks using a real robot.}
    \label{fig:Overview}
\end{figure*}

\subsection{Bin Packing Methods}
The vast majority of bin packing methods target industrial or warehouse domains. There, individual items are often pre-packed for transportation and their shape can be approximated by simple geometric priors. Thus, research has mainly focused on efficient bin space utilization or on object stability within the bin. 

Bin packing methods are commonly divided into online and offline settings.
Online bin packing methods process items in arrival order and place them sequentially without reordering.
GOPT~\cite{xiong2024gopt} follows this formulation and uses a Transformer architecture trained with reinforcement learning to place incoming items in sequence, while generalizing to different container sizes.
SDF-Pack~\cite{pan2023sdf} also operates online and optimizes packing compactness for objects of arbitrary shape represented as signed distance functions.

Offline bin packing methods assume that the full item set is known in advance, allowing the packing sequence to be optimized globally.
Jampacker~\cite{agarwal2020jampacker} targets cuboid warehouse items and minimizes free space while also introducing a learning-based module to correct imprecise placements.
For large-scale 3D bin packing, Elhedhli\etal~\cite{elhedhli2019} propose a layer-based column generation framework, whereas Dube \etal~\cite{dube2006} use a simulation-based heuristic that incrementally evaluates candidate pivot points with axis-aligned rotations.
Wang and Hauser~\cite{wang2019stable} incorporate both stability and execution feasibility for complex object geometries in a heuristic-based placement pipeline.

Only a few works address grocery packing, where the packing scheme is crucial for keeping the goods intact.
Chen\etal~\cite{chen2024real} propose a grocery packing system consisting of a proprioceptive soft gripper, tactile sensors, and \mbox{RGB-D} cameras. 
They leverage multimodal data to estimate the weight, size, and fragility of groceries and subsequently compute a packing score for each item. 
Similar to our work, Santos\etal~\cite{santos2024learning} propose to explicitly imitate human grocery packing preferences. 
To collect training data, human operators put a set of typical groceries into a box using a VR setup. 
A Markov chain model is then trained on the human-defined sequences. 
In contrast, our approach is zero-shot and open-vocabulary, i.e., it does not require training and can handle arbitrary objects. 

\subsection{Grocery Store Automation}
Several approaches to automate or simplify grocery shopping have been proposed. 
Cheng\etal~\cite{cheng2017design} present a shopping assistance robot capable of navigating through a supermarket. It autonomously picks items on the customer's shopping list from the shelves and places them into a basket. Contrary to our approach, it chooses the picking order to minimize travel distance rather than focusing on product integrity.
Thompson\etal~\cite{thompson2018autonomous} introduce a service robot for guiding customers to product locations in a grocery store. The robot also enables automatic billing and functions as an autonomous trolley, following a given shopping list.
Priya\etal~\cite{priya2021autonomous} follow a similar approach while adding an option for automatic payment with a connection to the customer's bank account.
Dworakowski\etal~\cite{dworakowski2021robot} present ContextSLAM, which enables autonomous navigation in dynamic grocery store environments and aids service robots in locating requested products. The aforementioned works rather focus on navigation, object detection, and object picking and do not consider the packing of the groceries. %

\section{The \algName\ approach}

Our goal is to pack a diverse set of items into a box while preserving product integrity and optimizing space utilization.
We achieve this by dividing the packing process into three major steps.
In the perception step, we detect and classify the objects in the camera image. Additionally, we also estimate the weight, size, and pose of each object.
In the planning step, we query an LLM to estimate packing constraints between the processed objects.
We then employ a mixed-integer optimization framework to compute a spatially optimal packing scheme given these constraints.
Lastly, we execute the suggested packing scheme with a robot arm using the execution module.
The overall method is summarized in Figure~\ref{fig:Overview}.

\subsection{Perception Step}
\label{sec:perception}

\begin{figure}
    \centering
    \includegraphics[width=\linewidth]{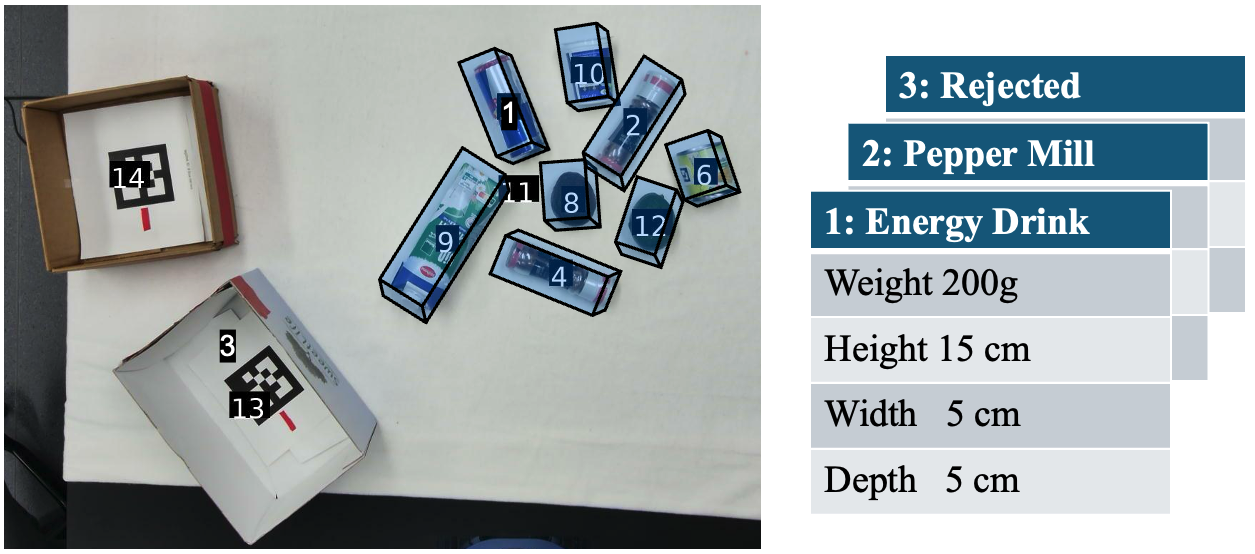}
    \caption{Visualization of the perception step. We detect objects using GroundingDINO. A VLM then infers the object type and properties.}
    \label{fig:vis_perception}
\end{figure}

First, we prompt GroundingDINO to identify all objects in the scene. 
SAM 2 is used in conjunction to obtain the respective segmentation masks.
While this step successfully identifies items, it is not robust to potential distractor objects in the image. 
Such objects include the box used for packing, the objects, or the robot's gripper.
We therefore add a VLM call for fine-grained object selection leveraging visual prompting~\cite{dettoolchain_wu_2024}. 
Specifically, we plot numbers in the middle of the bounding boxes identified by GroundingDINO and then ask the VLM to complete several tasks in a single inference (see Figure~\ref{fig:vis_perception}). The VLM is prompted to identify the subset of numbers that correspond to grocery items and to estimate the weight and size of each item. 

While it would be intuitive to estimate item dimensions from a point cloud, the measurements are subject to sensor noise and the object surface.
We therefore include the option for the VLM to access an agentic internet search where we provide the country of origin.
Since most grocery items come in standardized package sizes (e.g., beverage cans and bottles), an internet search can significantly improve the size and weight estimates.
Other non-standardized items, such as fruits and vegetables, are estimated by the VLM based on visual appearance.
To account for inaccuracies in size estimation, we add a collision padding in the subsequent planning step.
Objects on the table may move slightly when neighboring items are manipulated. To keep location estimates up to date, we track objects using the Hungarian algorithm with a cost that combines mask IoU and centroid distance. We also use mask IoU to verify successful grasps after each manipulation.

\subsection{Planning Step}
\label{sec:planning}

In the planning step, we determine the placement configuration of the detected objects in the target container. 
We consider a rectangular container with fixed dimensions $(W,D,H)$. 
Each item $i$ in the item list $I$ corresponds to an axis-aligned bounding box $(w_i, d_i, h_i)$ from the perception step and is modeled using variables for its Cartesian positions.
Constraints ensure that items can be rotated by $90^{\circ}$ around the main axes. %

Given the list $I$ of items recognized in the perception step, we prompt an LLM to find the set $\mathcal{F}$ of critical item pairs, i.e., $(i,j) \in \mathcal{F}$ iff placing $j$ on top of $i$ could damage item $i$. Next, we formulate item placement in the container as a linear programming problem that incorporates the above item safety constraints.
We further introduce contact avoidance constraints that prohibit item pairs from being placed side by side when contact could cause mutual damage, e.g., frozen spinach next to a bread loaf.
The solver then optimizes over the Cartesian item position. The following constraints are used:

\begin{enumerate}
    \item All items are fully placed in the container.
    \item The item bounding boxes do not overlap.
    \item For all $(i, j) \in \mathcal{F}$, if $i$ and $j$ overlap in the $x$-$y$ plane, then $i$ must be placed above $j$.
\end{enumerate}

The above procedure either returns a valid item placement or fails. This is sufficient when only one container size is available. 
In retail, however, the robot might have a choice between $K$ containers with dimensions $(W_k,D_k,H_k)$. 
In that case, we add all $K$ containers to the optimizer and include minimizing the container volume as an additional objective.
In the process, placements for non-chosen boxes are relaxed using a big M approach \cite{mip_book}.
A resulting packing scheme is visualized in Figure~\ref{fig:packingScheme}.

\begin{figure}
    \centering
    \input{planned_packing_binary.pgf}
    \caption{Example packing scheme returned by the optimizer. The bread loaf (brown) and the frozen spinach (green) are separated by contact avoidance constraints. The set $\mathcal{F}$ of critical item pairs ensures a protected position of the banana (yellow).}
    \label{fig:packingScheme}
\end{figure}

While many existing bin packing methods optimize for space utilization, only few consider the manipulation constraints imposed by real robotic systems.
Our planner can optionally incorporate them to ensure execution feasibility.
When enabled, three additional constraint types are added:
First, we query the LLM to identify geometrically unstable stacking configurations (e.g., cylindrical objects stacked directly on top of each other) and extend the constraint set $\mathcal{F}$ accordingly.
Second, to account for gripper dimensions and perception uncertainties, we apply \SI{5}{mm} collision padding around each object in the optimization.
Third, rather than permitting arbitrary rotations, we restrict objects to their most stable orientation---typically with the shortest dimension vertical---which simplifies manipulation.
To achieve stable packings, we finally minimize the summed vertical positions (z) of all items and solve with a \SI{30}{s} time limit, returning the best incumbent solution found.

\subsection{Execution Step}

\begin{figure}
    \centering
    \includegraphics[width=0.9\linewidth]{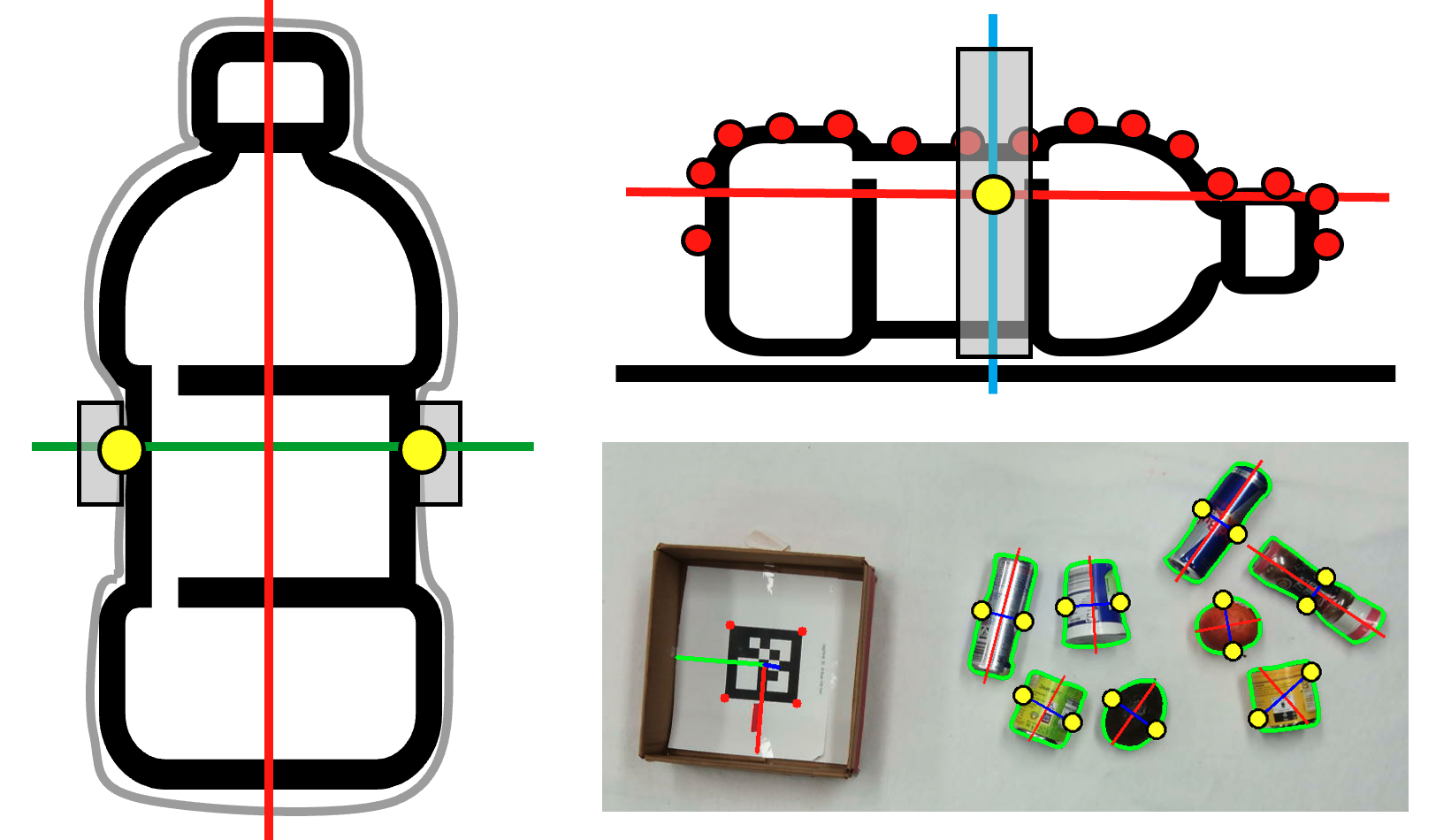}
    \caption{Visualization of the grasp pose construction. The grasping angle is aligned with the smaller principal component of the segmentation mask (left). The grasp position is the centroid of the RGB mask, with the $z$-coordinate set to the average of the RGB-D depth values (top-right). The approach direction is set to be vertical. We display the point cloud samples in red and the gripper-object contact points in yellow.}
    \label{fig:grasp_pose_construction}
\end{figure}

In the execution step, we move items to their target positions in the box according to the computed packing scheme.
Items are manipulated in ascending order by their z-positions. Thus, the lowest item is placed first.
The manipulation approach is agnostic to the choice of grasping and motion planning method; any planner capable of achieving high success rates with the required number of objects and sizes can be integrated.
While many modern AI-based grasping and manipulation methods exist, preliminary experiments showed that their success rates are insufficient for our use case.
For example, Octo~\cite{octo} and Open-VLA~\cite{openvla} successfully grasp known objects but fail to generalize to new objects without fine-tuning, thus violating the open-vocabulary requirement of our approach.
Contact Grasp Net~\cite{sundermeyer2021cgn}, on the other hand, successfully samples grasping poses from point clouds. 
However, the accuracy of the required point cloud largely depends on the object's surface material. 
While cardboard and vegetables yield accurate surface reconstructions, accuracy diminishes when dealing with translucent or reflective materials such as glass or aluminum.
To resolve these issues, we developed the following approach, which is based on a principal component analysis of the previously computed SAM 2 segmentation masks.
We project the achieved mask onto the scene point cloud to retrieve the points for the respective object and construct the grasp pose according to the following constraints (see Figure~\ref{fig:grasp_pose_construction}):
\begin{itemize}
    \item The approach direction is vertical (i.e., aligned with the global $z$-axis).
    \item The gripper fingers align with the smaller principal component of the segmentation mask.
    \item The grasp position is placed at the centroid of the RGB mask with the z coordinate being the average of the RGB-D depth values.
\end{itemize}

Once the grasp pose is computed, we execute the pick-and-place motion using a standard RRT motion planner. 
The robot lifts the object vertically, moves to a pre-placement pose above the target position specified by the packing scheme, and lowers the object to its final resting position. 
This sequence is repeated for each item in ascending order of $z$-position until the container is fully packed.

\section{Experimental Evaluation}
In this section, we experimentally validate our approach in simulation as well as using a real robot. In particular, we want to answer the following questions:
\begin{enumerate}
    \item How does the VLM choice influence the planning performance (Sec.~\ref{sec:sensitivity})?
    \item How does our planning approach compare with baseline methods that do not consider integrity constraints (Sec.~\ref{sec:baseline})?
    \item Are the integrity constraints preserved given the limitations of a real robot manipulator (Sec.~\ref{sec:full-stack})?
\end{enumerate}
Alongside, we introduce our validation metrics, dataset, and hardware setup.

\subsection{Dataset}
\label{sec:dataset}
We evaluate the planning performance of our method in a simulated environment.
Therefore, we require 3D meshes of common grocery items.
The HOPE-Dataset~\cite{lin2021fusion} is commonly used for such applications and consists of 3D scans of 28 widely available toy grocery items.
Although this serves as an excellent starting point, only a few of these items are fragile.
Therefore, we create \num{5} additional 3D scans of fragile objects and augment this dataset.
For the real-world experiments, we use a separate set of common household items acquired from a local grocery store. A subset is on display in Figure~\ref{fig:grasp_pose_construction}.

\subsection{Hardware Setup}
\label{section:hardware}
\begin{figure}[t]
  \centering
  \includegraphics[width=0.8\linewidth]{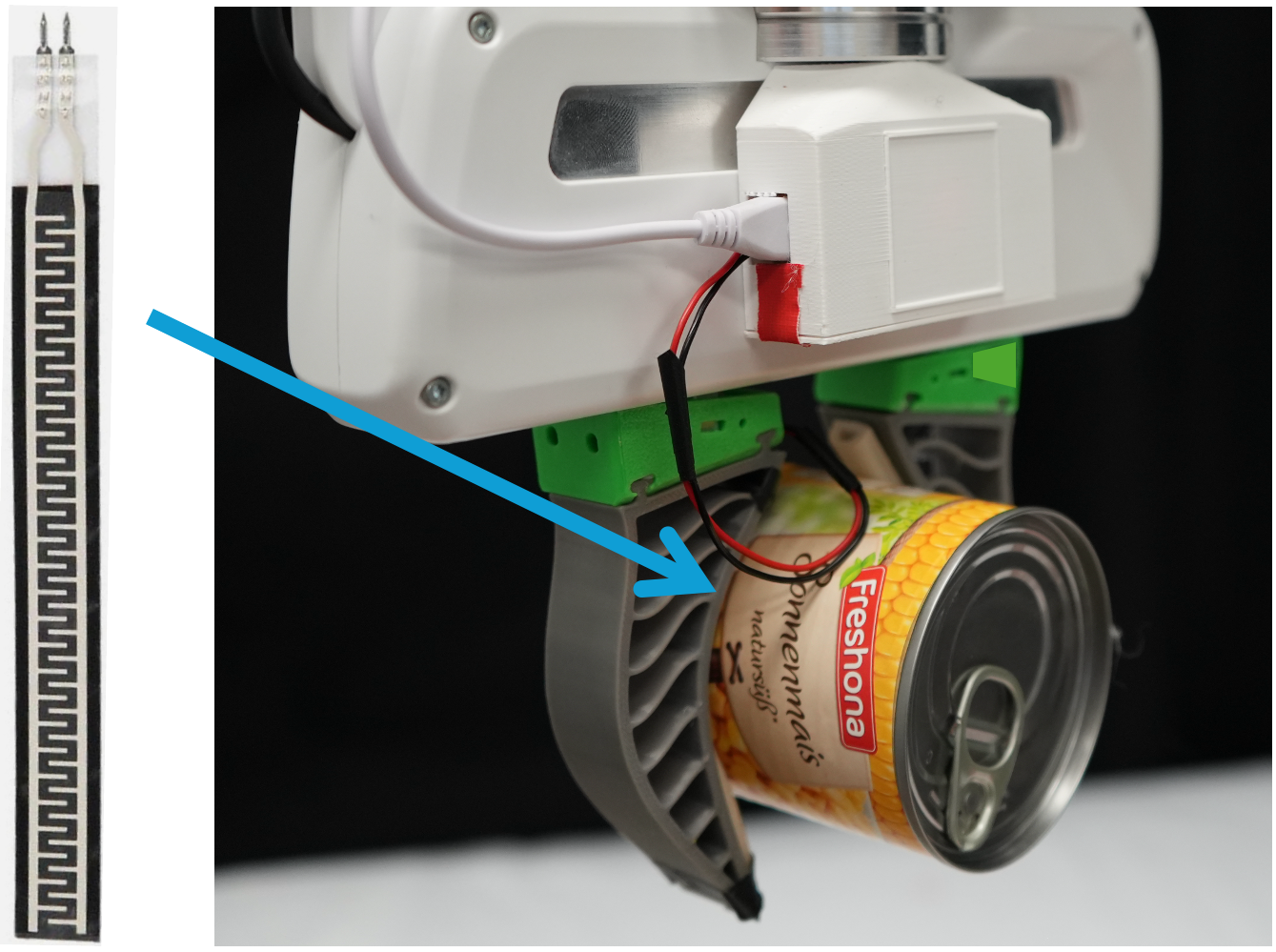}
  \caption{Visualization of our custom gripper. A thin-film force sensor is located in the left finger.}
  \label{fig:gripper}
\end{figure}

For real-world experiments, we employ a Franka Research~3 robot manipulator with custom 3D-printed gripper fingers that compliantly adapt to fragile items.
The gripper is based on the Fin Ray effect~\cite{pfaff2011application} and features an embedded force sensor (see Figure~\ref{fig:gripper}).
This enables closed-loop force control to accommodate objects of different weights and fragilities.
To capture \mbox{RGB-D} images, we deploy an Orbbec Femto Bolt camera.
In all experiments, we run model inference on a workstation with two NVIDIA RTX A6000 GPUs. 
The robot control code is executed by a separate computer running Debian Linux with a real-time kernel.

\subsection{\metricNameFull, a data-driven metric for human-like packing}
\label{sec:metric}

To quantify and compare the performance of different approaches to bin packing, we introduce a corresponding evaluation metric based on human packing preferences. Assume a set of obejct classes
\begin{equation*}
G= \{c_1, \dots, c_n\}.
\end{equation*}

\begin{figure}
    \centering
    \includegraphics[width=0.9\linewidth]{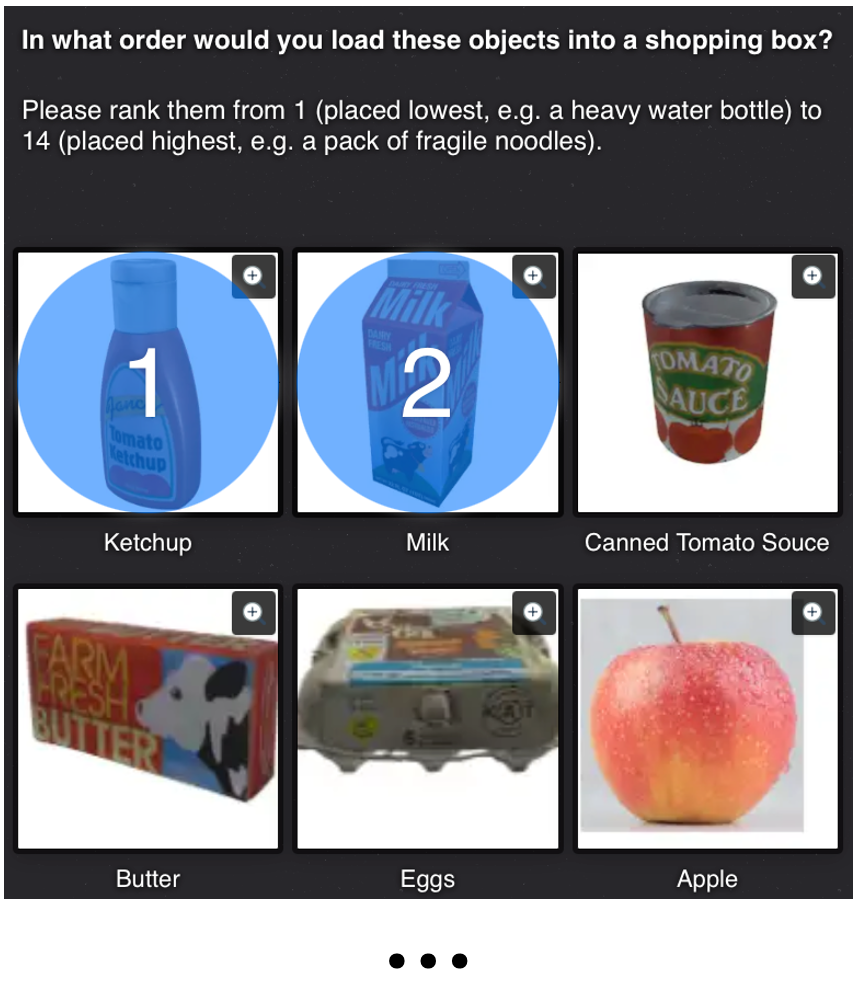}
    \caption{Visualization of the web-based survey. Participants are asked to click on the objects in their preferred packing sequence. The example items (heavy water bottle and fragile noodles) are not part of the actual survey to avoid biasing the results.}
    \label{fig:poll}
\end{figure}

We introduce a stacking preference for all classes as a binary relation $\leq$ on $G$.
For instance, $\textit{Canned Beans} \leq \textit{Eggs}$ means that \textit{Canned Beans} are preferred to be placed below \textit{Eggs}. Through a web-based poll, shown in Figure~\ref{fig:poll}, we empirically estimate the probability $P(c_i \leq c_k)$ of humans placing object class $c_i$ below $c_k$. 
The poll comprised 8,820 proposed object pairs across 14 object classes and was completed by 84 participants.

We evaluate a packing scheme by counting violations of human stacking preferences that occur when two objects geometrically overlap in the box. We form the set $\mathcal{O}$ of ordered object class pairs $(i,k)$ for which the projections of the placed meshes onto the $x$--$y$ plane overlap and object $k$ is placed above object $i$.
We use the human-derived probabilities $P(c_i \leq c_k)$ to decide whether a stacked pair constitutes a violation. We count a pair $(i,k)\in\mathcal{O}$ as a violated constraint if this stacking relation is unlikely under the human data:
\[
\mathbb{I}_{\text{viol}}(i,k)=
\begin{cases}
1, & \text{if } P(c_i \leq c_k) < \theta\\
0, & \text{otherwise.}
\end{cases}
\]
We use the threshold $\theta$ (0.4 in our experiments) to reduce the sensitivity to noisy or ambiguous human annotations by penalizing only relations that are clearly disfavored.
Finally, we define the \metricNameFull~of each box as the total number of constraint violations:
\[
C=\sum_{(i,k)\in\mathcal{O}} \mathbb{I}_{\text{viol}}(i,k).
\]

We evaluate the performance of our method based on the average number of violated constraints per box across scenes ($aC$), the achieved packing density, and the required time per box.
The packing density is the ratio between the summed object volumes and the volume of the box.

\subsection{VLM Sensitivity Study}
\label{sec:sensitivity}
We evaluate our method's ability to propose a human-like packing scheme in a standalone manner. 
For each trial, we randomly select four fragile and five non-fragile objects from our dataset. We initialize the evaluation with a box of edge length \SI{50}{cm}, which is sufficiently large for every combination of items. 
We then perform a binary search on the box dimensions to identify the minimal box size up to a precision of \SI{2}{mm}.
In the process, we ensure that the box remains large enough to fit any object in a stand-alone setting. 
This way, we avoid premature termination of the shrinking process caused by large single items.
The procedure is repeated ten times to obtain statistically viable information. 
We find that API response times exhibit high variance and thus report the optimization time without API delay.

\begin{table}[h]
\centering
\caption{Results of the VLM sensitivity study in terms of density, $aC$ and planning time. $N_o$ denotes the average number of item overlaps. Optimization times do not include the response delay of the API.}
\label{tab:Sensitivity}
\tabsensitivityStudy
\end{table}

The result is on display in Table~\ref{tab:Sensitivity}.
For the generation of our product integrity constraints, we first evaluate \num{4} VLMs, specifically GPT-5~\cite{singh2025openaigpt5card}, GPT-5-mini~\cite{singh2025openaigpt5card}, Claude Sonnet 4.5~\cite{anthropic_claude_sonnet_4_5}, and Gemini 3 Flash \cite{gemini2023}.
Claude Sonnet 4.5 scores the lowest in terms of \metricName~and best in terms of density.
This stems from generating a comparably small set of constraints that does not align well with human preferences, leading to less restriction on the underlying MIP approach and ultimately resulting in higher density.
GPT-5, on the other hand, delivers the best result at 0.2~\metricName. GPT-5 is therefore the VLM used in the following experiments.

\subsection{Planning Baseline Comparison} 
\label{sec:baseline}
We compare against the offline methods of Elhedhli\etal~\cite{elhedhli2019} and Dube \etal~\cite{dube2006}\footnote{We use the public implementations available at \url{github.com/Wadaboa/3d-bpp} \cite{elhedhli2019} and \url{github.com/jerry800416/3D-bin-packing} \cite{dube2006}}, neither of which incorporates product integrity constraints, as well as the online method GOPT~\cite{xiong2024gopt}.
As an online bin packing method, GOPT has no information about the types and sizes of downstream items.

\begin{table}[t]
     \centering
     \caption{Results of the packing evaluation in terms of density and $aC$. \algName~(vanilla) denotes our approach without product integrity constraints. $N_o$ denotes the average number of item overlaps.}
     \label{tab:Baseline}
     \tabbaseline
\end{table}

Table~\ref{tab:Baseline} summarizes the results. Although the baseline by Dube\etal\ is the fastest, \algName~is the best with respect to violated constraints and packing density. The highest density is achieved by \algName~without enforced integrity constraints, whereas \algName~achieves the lowest number of violated constraints.
It should be noted that the runtime of our optimization approach is not constant but rather depends on the complexity of the problem. 
Thus, it increases when the density approaches the theoretical optimum. 

The time reported in Table~\ref{tab:Baseline} thus approaches the worst-case scenario. In practice, we apply a time limit of \SI{30}{s} per optimization run, after which the best solution found so far is returned or the process is stopped.

\subsection{Full System Evaluation} 
\label{sec:full-stack}
\begin{figure}
    \centering
    \includegraphics[width=0.9\linewidth]{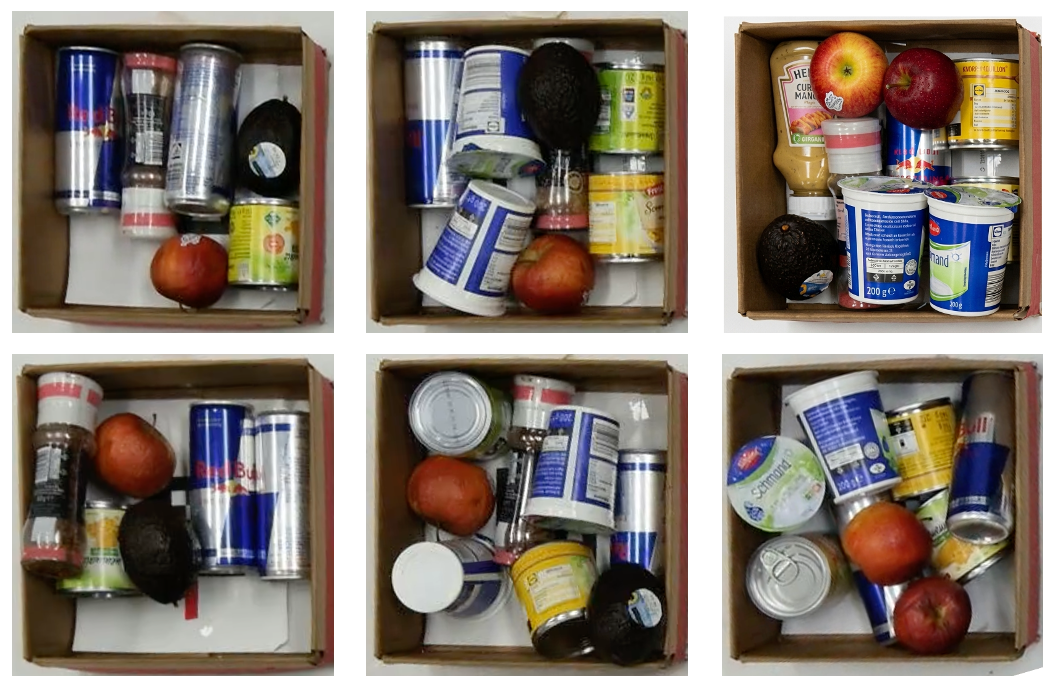}
    \caption{Results of the Full System Evaluation. We show packed boxes for \num{6} (left), \num{8} (middle) and \num{10} (right) objects.}
    \label{fig:packed_boxes}
\end{figure}

To validate the practical applicability of \algName, we conduct end-to-end experiments in a real-world setting. 
Our experimental setup requires several assumptions about the initial object configuration: objects must lie flat (with their shortest principal component pointing upward) at both source and target poses, fit into the gripper at their centroid, and not overlap on the table.
A simulated benchmark indicates that these assumptions, together with the hardware-specific constraints introduced in Section~\ref{sec:planning}, reduce the average packing density to \SI{26.4}{\%}.
We pack item sets of 6, 8, 10, and 12 objects and perform 5 trials per size with different objects and initial configurations (see Figure \ref{fig:packed_boxes}).
Here, the algorithm selects from a set of two available packing containers. 
We evaluate the achieved \metricName~score and manipulation success rate.
A manipulation is successful if an object is picked from the table and placed in the box.
Figure~\ref{fig:realWorldScaling} shows that \metricName~increases with the number of processed items, indicating more constraint violations in larger item sets as expected from the planning evaluation.

\begin{figure}
    \centering
    \input{FullSystemScaling.pgf}
    \caption{Results of the real-world experiments. We report the C metric and the manipulation success rate.}
    \label{fig:realWorldScaling}
\end{figure}

Across all trials, the approach achieves a manipulation success rate of \SI{92.6}{\%}. 
Failure cases can be attributed to three primary sources: 
First, noise in the \mbox{RGB-D} data propagates to the grasp pose estimation, causing insufficient grip stability and object slippage during manipulation.
Second, the gripper dimensions cause frequent collisions with the container walls during placement, a problem that is more pronounced in larger boxes where more objects can be placed near the walls.
Third, inaccuracies in placing the object lead to object-object collisions and constraint violations in downstream manipulations.
These limitations indicate several directions for future work. 
For instance, VLA-based manipulation policies could perform contact-rich behaviors such as pushing, providing additional degrees of freedom for object placement.
Such approaches would also enable closed-loop error correction when placement deviations or manipulation errors are detected.

\section{Conclusions}
In this paper, we introduced \algName, a novel approach to open-vocabulary object packing that ensures product integrity. To achieve its task, \algName{} utilizes Vision Language Models and state-of-the-art Mixed-Integer optimization. 
We validated our approach on the HOPE dataset, extended by several new grocery classes. 
We quantified the performance of \algName~based on packing density and on a custom-developed data-driven metric \metricName. 
The experiments demonstrate that \algName~shows great performance in packing objects according to human-preferred packing order. One opportunity for future work is to increase the robustness of the placement process by applying specialized vision-language-action models that leverage pushing motions to correct misaligned placements.

\section*{Acknowledgment}
This work was partially supported by the German Federal Ministry of Research, Technology and Space (BMFTR) under the Robotics Institute Germany (RIG).

\bibliographystyle{IEEEtran}
\bibliography{sources.bib}

\begin{thebibliography}{10}
\providecommand{\url}[1]{#1}
\csname url@samestyle\endcsname
\providecommand{\newblock}{\relax}
\providecommand{\bibinfo}[2]{#2}
\providecommand{\BIBentrySTDinterwordspacing}{\spaceskip=0pt\relax}
\providecommand{\BIBentryALTinterwordstretchfactor}{4}
\providecommand{\BIBentryALTinterwordspacing}{\spaceskip=\fontdimen2\font plus
\BIBentryALTinterwordstretchfactor\fontdimen3\font minus \fontdimen4\font\relax}
\providecommand{\BIBforeignlanguage}[2]{{%
\expandafter\ifx\csname l@#1\endcsname\relax
\typeout{** WARNING: IEEEtran.bst: No hyphenation pattern has been}%
\typeout{** loaded for the language `#1'. Using the pattern for}%
\typeout{** the default language instead.}%
\else
\language=\csname l@#1\endcsname
\fi
#2}}
\providecommand{\BIBdecl}{\relax}
\BIBdecl

\bibitem{cordeiro2022bin}
A.~Cordeiro, L.~F. Rocha, C.~Costa, P.~Costa, and M.~F. Silva, ``Bin picking approaches based on deep learning techniques: A state-of-the-art survey,'' in \emph{IEEE Int. Conf. on Autonomous Robot Systems and Competitions (ICARSC)}, 2022.

\bibitem{pantoja2024comprehensive}
G.~Pantoja-Benavides, D.~Giraldo, A.~Montes \emph{et~al.}, ``Comprehensive review of robotized freight packing,'' \emph{Logistics}, vol.~8, no.~3, 2024.

\bibitem{amoo2024warehouse}
O.~Amoo, E.~Sodiya, U.~Umoga, and A.~Atadoga, ``Ai-driven warehouse automation: A comprehensive review of systems,'' \emph{GSC Advanced Research and Reviews}, vol.~18, pp. 272--282, 2024.

\bibitem{agarwal2020jampacker}
M.~Agarwal, S.~Biswas, C.~Sarkar, S.~Paul, and H.~S. Paul, ``Jampacker: An efficient and reliable robotic bin packing system for cuboid objects,'' \emph{IEEE Robotics and Automation Letters (RA-L)}, 2020.

\bibitem{wang2019stable}
F.~Wang and K.~Hauser, ``Stable bin packing of non-convex {3D} objects with a robot manipulator,'' in \emph{IEEE Int.~Conf.~on Robotics and Automation (ICRA)}, 2019.

\bibitem{wang2022dense}
------, ``Dense robotic packing of irregular and novel 3d objects,'' \emph{IEEE Transactions on Robotics (T-RO)}, 2022.

\bibitem{cheng2017design}
C.-H. Cheng, C.-Y. Chen, J.-J. Liang, T.-N. Tsai, C.-Y. Liu, and T.-H.~S. Li, ``Design and implementation of prototype service robot for shopping in a supermarket,'' in \emph{Int.~Conf.~on Advanced Robotics and Intelligent Systems (ARIS)}, 2017.

\bibitem{thompson2018autonomous}
C.~Thompson, H.~Khan, D.~Dworakowski, K.~Harrigan, and G.~Nejat, ``An autonomous shopping assistance robot for grocery stores,'' in \emph{Workshop on Robotic Co-workers}, 2018.

\bibitem{priya2021autonomous}
S.~N. Priya, G.~Swadesh, K.~Thirivikraman, M.~V. Ali, and M.~R. Kumar, ``Autonomous supermarket robot assistance using machine learning,'' in \emph{Int.~Conf.~on Advanced Computing and Communication Systems (ICACCS)}, 2021.

\bibitem{dworakowski2021robot}
D.~Dworakowski, C.~Thompson, M.~Pham-Hung, and G.~Nejat, ``A robot architecture using contextslam to find products in unknown crowded retail environments,'' \emph{Robotics}, vol.~10, no.~4, p. 110, 2021.

\bibitem{liu2024grounding}
S.~Liu, Z.~Zeng, T.~Ren \emph{et~al.}, ``Grounding dino: Marrying dino with grounded pre-training for open-set object detection,'' in \emph{Europ.~Conf.~on Computer Vision (ECCV)}, 2024.

\bibitem{ravi2025sam}
N.~Ravi, V.~Gabeur, Y.-T. Hu \emph{et~al.}, ``{SAM} 2: Segment anything in images and videos,'' in \emph{The Thirteenth International Conference on Learning Representations}, 2025.

\bibitem{Maher2016}
S.~Maher, M.~Miltenberger, J.~P. Pedroso, D.~Rehfeldt, R.~Schwarz, and F.~Serrano, ``{PySCIPOpt}: Mathematical programming in {Python} with the {SCIP} optimization suite,'' in \emph{Mathematical Software (ICMS)}, 2016.

\bibitem{xiong2024gopt}
H.~Xiong, C.~Guo, J.~Peng \emph{et~al.}, ``{GOPT}: Generalizable online 3d bin packing via transformer-based deep reinforcement learning,'' \emph{IEEE Robotics and Automation Letters (RA-L)}, 2024.

\bibitem{pan2023sdf}
J.-H. Pan, K.-H. Hui, X.~Gao \emph{et~al.}, ``Sdf-pack: Towards compact bin packing with signed-distance-field minimization,'' in \emph{IEEE/RSJ Int.~Conf.~on Intelligent Robots and Systems (IROS)}, 2023.

\bibitem{elhedhli2019}
S.~Elhedhli, F.~Gzara, and B.~Yildiz, ``Three-dimensional bin packing and mixed-case palletization,'' \emph{INFORMS Journal on Optimization}, 2019.

\bibitem{dube2006}
E.~Dube and L.~Kanavathy, ``Optimizing three-dimensional bin packing through simulation,'' in \emph{Proc. of the Int. Conf. on Computational Modelling, Simulation and Optimization (ICCMSO)}, 2006.

\bibitem{chen2024real}
V.~K. Chen, L.~Chin, J.~Choi, A.~Zhang, and D.~Rus, ``Real-time grocery packing by integrating vision, tactile sensing, and soft fingers,'' in \emph{Int.~Conf.~on Soft Robotics (RoboSoft)}, 2024.

\bibitem{santos2024learning}
A.~Santos, N.~F. Duarte, A.~Dehban, and J.~Santos-Victor, ``Learning the sequence of packing irregular objects from human demonstrations: Towards autonomous packing robots,'' in \emph{Int.~Conf.~for Biomedical Robotics and Biomechatronics (BioRob)}, 2024.

\bibitem{dettoolchain_wu_2024}
Y.~Wu, Y.~Wang, S.~Tang \emph{et~al.}, ``{DetToolChain}: A new prompting paradigm to unleash detection ability of {MLLM},'' in \emph{Europ.~Conf.~on Computer Vision (ECCV)}, 2024.

\bibitem{mip_book}
J.~M. Giron-Sierra, ``Integer programming,'' in \emph{Introduction to Optimization with Matlab Examples}.\hskip 1em plus 0.5em minus 0.4em\relax Springer Nature Switzerland, 2014.

\bibitem{octo}
D.~Ghosh, H.~Walke, K.~Pertsch \emph{et~al.}, ``{Octo}: An open-source generalist robot policy,'' in \emph{Robotics: Science and Systems (RSS)}, 2024.

\bibitem{openvla}
M.~J. Kim, K.~Pertsch, S.~Karamcheti \emph{et~al.}, ``{OpenVLA}: An open-source vision-language-action model,'' in \emph{Conf.~on Robot Learning (CoRL)}, 2024.

\bibitem{sundermeyer2021cgn}
M.~Sundermeyer, A.~Mousavian, R.~Triebel, and D.~Fox, ``Contact-graspnet: Efficient 6-dof grasp generation in cluttered scenes,'' in \emph{IEEE Int.~Conf.~on Robotics and Automation (ICRA)}, 2021.

\bibitem{lin2021fusion}
Y.~Lin, J.~Tremblay, S.~Tyree, P.~A. Vela, and S.~Birchfield, ``Multi-view fusion for multi-level robotic scene understanding,'' in \emph{IEEE/RSJ International Conference on Intelligent Robots and Systems (IROS)}, 2021, pp. 6817--6824.

\bibitem{pfaff2011application}
O.~Pfaff, S.~Simeonov, I.~Cirovic, P.~Stano \emph{et~al.}, ``Application of fin ray effect approach for production process automation,'' \emph{Annals of DAAAM \& Proceedings}, vol.~22, no.~1, pp. 1247--1249, 2011.

\bibitem{singh2025openaigpt5card}
A.~Singh, A.~Fry, A.~Perelman \emph{et~al.}, ``{OpenAI} {GPT-5} system card,'' \url{https://arxiv.org/abs/2601.03267}, 2025.

\bibitem{anthropic_claude_sonnet_4_5}
\BIBentryALTinterwordspacing
{Anthropic}. (2025) Claude sonnet 4.5. [Online]. Available: \url{https://www.anthropic.com/news/claude-sonnet-4-5}
\BIBentrySTDinterwordspacing

\bibitem{gemini2023}
{Gemini Team}, R.~Anil, S.~Borgeaud \emph{et~al.}, ``Gemini: A family of highly capable multimodal models,'' \url{https://arxiv.org/abs/2312.11805}, 2025.

\end{thebibliography}

\end{document}